\pdfoutput=1
\documentclass[runningheads]{llncs}
\usepackage{graphicx}
\usepackage{times}
\usepackage{latexsym}
\usepackage{soul}
\usepackage{amsmath}
\usepackage{multirow}
\usepackage{subcaption}
\usepackage{hyperref}
\usepackage{listings}
\usepackage{caption}

\usepackage[T1]{fontenc}
\usepackage{xcolor}
\usepackage[utf8]{inputenc}

\usepackage{microtype}

\usepackage{inconsolata}

\begin{document}
\title{KG-CTG: Citation Generation through Knowledge Graph-guided Large Language Models}
\titlerunning{KG-CTG}
\author{Avinash Anand\orcidID{0009-0003-2479-0342} \and
Mohit Gupta\orcidID{0009-0001-8528-5169} \and
Kritarth Prasad\orcidID{0009-0006-2279-7112}\and
Ujjwal Goel\orcidID{0009-0004-2025-9359} \and
Naman Lal\orcidID{0009-0008-2914-5509} \and
Astha Verma\orcidID{0000-0003-3615-5373} \and
Rajiv Ratn Shah\orcidID{0000-0003-1028-9373}
}
\author{Avinash Anand \and
Mohit Gupta \and
Kritarth Prasad \and
Ujjwal Goel \and
Naman Lal \and
Astha Verma \and
Rajiv Ratn Shah
}
\institute{Indraprastha Institute of Information Technology, Delhi \\
\email{\{avinasha, mohit22112, kritarth20384, ujjwal20545, asthav, rajivratn\}@iiitd.ac.in}\\
\email{namanlal.lal92@gmail.com}
}
\authorrunning{A. Anand et al.}
\maketitle
\begin{abstract}
Citation Text Generation (CTG) is a task in natural language processing (NLP) that aims to produce text that accurately cites or references a cited document within a source document. In CTG, the generated text draws upon contextual cues from both the source document and the cited paper, ensuring accurate and relevant citation information is provided. Previous work in the field of citation generation is mainly based on the text summarization of documents. Following this, this paper presents a framework, and a comparative study to demonstrate the use of Large Language Models (LLMs) for the task of citation generation. Also, we have shown the improvement in the results of citation generation by incorporating the knowledge graph relations of the papers in the prompt for the LLM to better learn the relationship between the papers. To assess how well our model is performing, we have used a subset of standard S2ORC dataset, which only consists of computer science academic research papers in the English Language. Vicuna performs best for this task with 14.15 Meteor, 12.88 Rouge-1, 1.52 Rouge-2, and 10.94 Rouge-L. Also, Alpaca performs best, and improves the performance by 36.98\% in Rouge-1, and 33.14\% in Meteor by including knowledge graphs. 
\end{abstract}
\section{Introduction}
% \subsubsection{WHAT IS CTG ?}
Generating text in the scientific domain is a complex task that demands a strong grasp of the input text and domain-specific knowledge. Citation Text Generation (CTG) is an NLP task that focuses on generating accurate citations or references to cited documents within a source document. To accomplish this, machine learning models must adeptly summarize the relationship between the original and the cited article in a given context. This involves analyzing the content of the documents, identifying their connections, and employing appropriate terminology and structure to convey this information with clarity and conciseness. The domain of Text Generation has gained significant attention in recent times, largely attributed to the advancements in Transformer-based models~\cite{vaswani2017attention}. These models have revolutionized the field, driving substantial progress in natural language processing and generation tasks.

CTG holds great potential for various applications, particularly in the field of scientific writing assistants. In the context of Education, CTG could be employed to teach students the correct approach to citing papers in academic writing. Another intriguing prospect is the ability to create summary sentences~\cite{agarwal2011scisumm} from the referenced source papers, effectively summarizing the key ideas of the cited text. Next significant use is assisting researchers in paper writing by providing suggestions for suitable citations and generating the corresponding citation text. It can aid in plagiarism detection by comparing the generated citation text with the source material. CTG is a complementary task to citation recommendation and summarization, as it specifically focuses on explaining the relationships between documents rather than just summarizing their contents ~\cite{cohan2017scientific,yasunaga2019scisummnet}.

\begin{figure}[ht]
\centering
  \includegraphics[width = \linewidth]{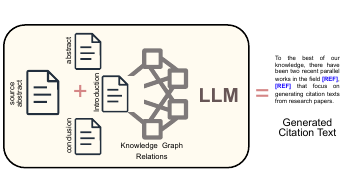}
  \caption{Single-Sentence Citation Text Generation}
\label{fig:single_text_citation_dataset}
\end{figure}

Scientific texts are significantly longer than other domains typically studied in NLP. This is a challenging and unresolved problem for text generation models. This problem was solved by a few works like, Xing et al.~\cite{xing2020automatic} does automatic generation of citation text in scholarly articles. It makes use of a cross-attention multi-source pointer generating network. Luu et al.~\cite{Luu2020CitationTG} introduced CTG using a pair of scientific documents as the source and cited papers. They have primarily used the text from the abstract of papers as input. It generated the citation text describing the relationship between the source and cited documents. 

% \begin{figure}[ht]
% \centering
%   \includegraphics[width = \linewidth]{emnlp2023-latex/single_ref_citations.pdf}
%   \caption{Example of Single Citation Text Generation given citing and cited paper abstracts.}
% \label{fig:citation_example}
% \end{figure}

To our knowledge, researchers use subsets of the S2ORC\footnote{\url{https://github.com/allenai/s2orc}}~\cite{lo2019s2orc} dataset for the CTG task. We have only used the subset which contains research articles only in the domain of computer science in English Language. The original dataset contains $abstract$ of paper, $body\_text$, $paper\_id$, etc. We have extracted the \textbf{introduction}, and \textbf{conclusion} from the $body\_text$. 

In this paper, we proposed a methodology or technique to generate citation text using Large Language Models (LLMs). We have fine-tuned three LLMs for the task of generating citation text. The models are LLaMA~\cite{LLaMA}, Alpaca~\cite{alpaca}, and Vicuna~\cite{vicuna2023}. We have also performed the experiments of incorporating the Knowlegdge graph in the prompts for generating the citation, to provide better contextual understanding of the source and target paper to better learn the relationship between them. Incorporating the knowledge graphs shows the improvement in the performance and quality of text generation. To conclude, our main contributions in this paper is listed below:

\begin{itemize}
\item We propose the use of Large Language Model (LLMs) text generation capabilities in field of research writing. We fine-tuned three LLMs for generating the citation text given the content of citing and cited papers.
\item We attempt to incorporate the knowledge graphs of the source and target papers in the prompts to make the model better understand the relationship between the papers. For creating the knowledge graphs, we have used \textbf{PL-Marker}~\footnote{\url{https://github.com/thunlp/PL-Marker}}~\cite{ye2022plmarker}. 
\item We show the importance of incorporating the knowledge graphs to improve the performance. We achieve an increase of 33.14\% in METEOR and 36.98\% in Rouge-1 score for Alpaca on the S2ORC dataset.
\end{itemize}

The structure of the written work is as follows: Section \ref{literature} provides an overview of the related works on citation text generation. Section \ref{methodology} describes the problem formulation, the utilized models, and their components. Section \ref{dataset} outlines the processing and creation of the dataset from S2ORC. Section \ref{experiments} presents the experimental setup, findings, and implementation details. Section \ref{evaluations} presents the evaluations conducted in the study. Section \ref{conclusion} summarizes the future aims of the paper. Lastly, Section \ref{limitations} discusses the limitations of the proposed system.

\section{Related Work\label{literature}}
CTG is intricately connected to citation recommendation, scientific document understanding, and summarization. The task of citation recommendation complements CTG, as it provides references to pertinent publications for a particular document or text excerpt~\cite{bornmann2015growth}. Additionally, citation recommendation systems~\cite{bhagavatula2018content} play a vital role in guiding researchers towards valuable sources of information. Summarization Systems~\cite{yasunaga2019scisummnet} condense the information allowing scientists to understand the basic idea in a research section more quickly. 

Citation information is also helpful for scientific paper summarization~\cite{qazvinian2008scientific}. Previous work include the works of \cite{jaidka2010imitating} in their literature review producing system for text summarization. The task of multi-document summarization in the scientific domain~\cite{chen2014summarization} and text generation for scientific documents is a particular case of multi-document scientific summarization~\cite{chen2019automatic}. 

Koncel-Kedziorski et al.~\cite{koncel2019text} generated multi-sentence text from an information extraction system and improved performance using a knowledge graph. They did graph encoding using Graph Attention Network. Chen et al.~\cite{chen2021scixgen}~proposed a SciXGen dataset to solve the problem of context-aware text generation in a scientific domain. Zhu et al.~\cite{zhu2023llms} uses LLMs for constructing the knowledge graphs, and used them for reasoning. They proposed \textbf{AutoKG}, which uses LLMs for constructing and reasoning the knowledge graphs. Gosangi et al.~\cite{gosangi2021use} studied the significance of context in determining whether a sentence in an academic publication is worthy of reference. This paper can be considered complementary work with CTG tasks. 
% ~\cite{Luu2020CitationTG}\cite{ xing2020automatic}

To the best of our knowledge, there have been two recent parallel works in the field that focus on generating citation texts from research papers. Luu et al.~\cite{Luu2020CitationTG} were the first to introduce this task and successfully generated citation texts using the source and cited documents as input. On the other hand, Xing et al.~\cite{xing2020automatic} delved deeper into the relationship between scientific documents by leveraging a larger dataset. They employed an implicit citation extraction algorithm, utilizing GPT-2\footnote{\url{https://github.com/openai/gpt-2}}, which was trained on an annotated dataset to automatically enhance the training data. The rise of large language models (LLMs) has brought about a major breakthrough in education~\cite{10.1007/978-3-031-49601-1_5,10.1007/978-3-031-49601-1_4,mathify}, and also in CTG tasks, as shown by recent research like that of Avinash et al.~\cite{inbook_context_enhanced}. These flexible models have created many new learning possibilities.

\section{Methodology\label{methodology}}
The goal of the task of citation generation, which tries to produce citation text in the context of both the source publication and the referenced paper, is outlined in this section. We leverage the advancements of Large Language Models (LLMs), which have demonstrated significant improvements in various text generation tasks~\cite{zhao2023survey}.
\begin{itemize}
    \item \textbf{Model Fine-Tuning: } We commence by fine-tuning three selected LLMs i.e. LLaMA~\cite{LLaMA}, Alpaca~\cite{alpaca}, and Vicuna~\cite{vicuna2023} specifically for the task of generating citation text. Through this process, we train the models on the subset of S2ORC dataset for citation generation and evaluate their performance and the quality of the generated text.
    \item \textbf{Incorporation of Knowledge Graphs: } To improve the model's comprehension of the context and relationship between the papers, we introduce a knowledge graph~\cite{koncel2019text} derived from the papers as part of the input prompt. The knowledge graph provides structured information about the papers, including key concepts, entities, and their relationships. We explore the benefits of incorporating knowledge graphs with LLMs~\cite{pan2023unifying} in improving the performance of citation generation tasks.
\end{itemize}

By following this methodology, we show the effectiveness of fine-tuned LLMs for generating citation text. Additionally, we look into the effect of incorporating knowledge graphs on improving the model's performance in capturing the relationships and rich contextual information between the source paper and the cited paper. 

\begin{figure*}[ht]
\centering
  \includegraphics[width=\linewidth]{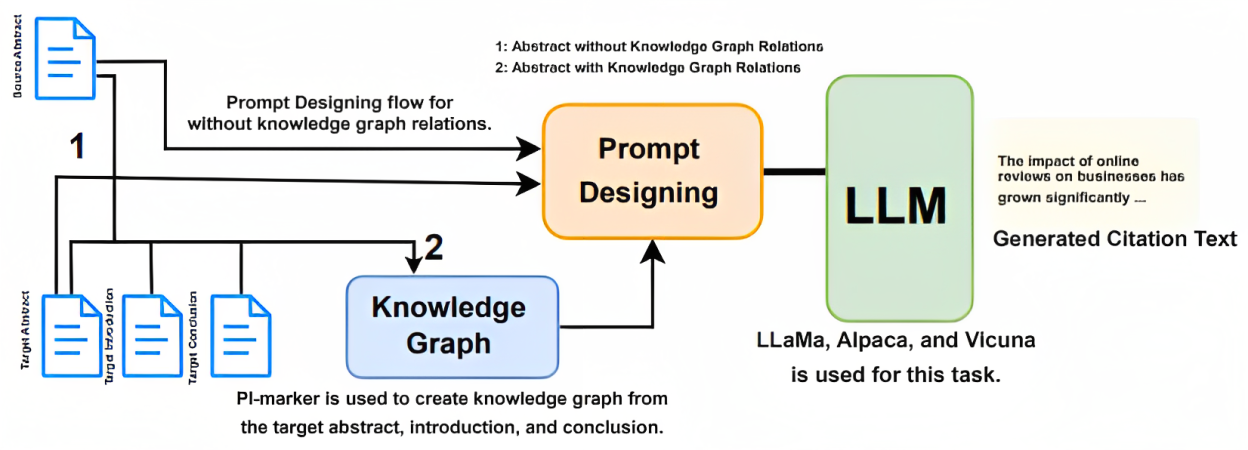}
  \caption{Experimental setup workflow diagram depicting two workflows: \textbf{1}. Prompt creation without knowledge graph relations, and \textbf{2}. Relation extraction from abstracts followed by prompt creation.} 
\label{fig:fig_2}
\end{figure*}
\subsection{\textbf{Large Language Models}}
In this subsection, we describe the Large Language Models (LLMs) used in our study for citation text generation. The LLMs investigated in our research include LLaMA~\cite{LLaMA}, Alpaca~\cite{alpaca}, and Vicuna~\cite{vicuna2023}. LLaMA is a transformer-based model available in multiple variations, such as 7B, 13B, 33B, and 65B parameters. For our study, we focused on LLaMA-7B. Alpaca, a variant of LLaMA, has been fine-tuned using 52k instructions from OpenAI's text-davinci-003 model. This targeted training allows Alpaca to specialize in generating instructional text. Vicuna is a supervised fine-tuned version of LLaMA, trained on 70K user-shared talks from ShareGPT.com. This variant of LLaMA captures the specific context and style of user-shared talks, enabling it to generate text aligned with conversational patterns.
% \begin{equation}
% \begin{split}
%     \small {\#\textbf{input}: source\_abstract}\\
%     \small {target\_abstract}\\
%     \small {\#\textbf{response}: citation\_text}
% \label{eq: equation1}
% \end{split}
% \end{equation}

To evaluate the performance of these LLMs on the task of citation text generation (CTG), we fine-tuned all models using our CTG dataset. This comparative analysis enables us to assess the strengths and weaknesses of each model in generating accurate and contextually appropriate citation text.

\subsection{\textbf{Knowledge Graphs \& Prompting}}
The structured representation of knowledge is called knowledge graph that organizes information into entities and their relationships, enabling advanced data analysis and inference~\cite{Ji_2022}. In our work, we constructed the knowledge graph of the source and target abstracts using a state-of-the-art tool called PL-Marker~\cite{ye2022plmarker}. PL-Marker employs a novel packed levitated marker technique, combining a packing approach focused on the subject and the neighbourhood to obtain pair representations. The knowledge graph is constructed to capture the relationship and context between different entities within the abstracts of papers. 

\begin{figure}
\centering
\begin{subfigure}{.5\textwidth}
\centering  \includegraphics[width=.7\linewidth]{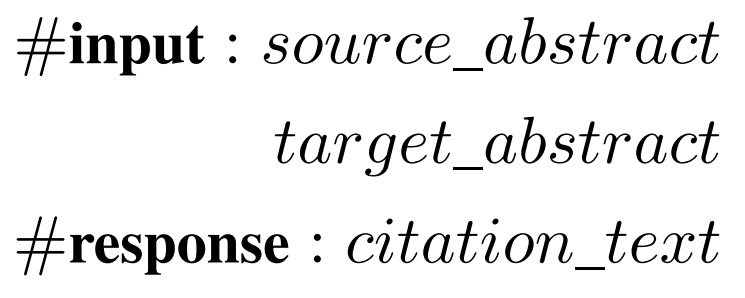}
  \caption{}
  \label{eq: equation1}
\end{subfigure}%
\begin{subfigure}{.5\textwidth}
  \centering
\includegraphics[width=.8\linewidth]{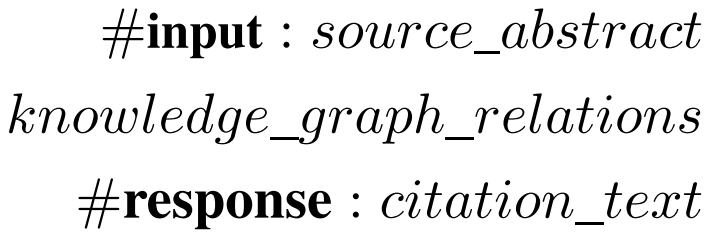}
  \caption{}
  \label{eq: equation2}
\end{subfigure}
\caption{Prompt Structures used for the Large Language Models.}
\label{Prompts}
\end{figure}

After generating the knowledge graph from the source abstract with the target abstract, introduction and conclusion, we then concatenated it with the source abstract. This final string is then passes in the $\#input$ section of the prompt. The structure of the prompt with knowledge graph relations is shown in~\ref{eq: equation2}.
% \begin{equation}
% \begin{split}
%     \small {\#\textbf{input}: source\_abstract\\
%     knowledge\_graph\_relations\\
%     \#\textbf{response}: citation\_text}
% \label{eq: equation2}
% \end{split}
% \end{equation}

\section{CTG Dataset\label{dataset}}

In our research, we utilized the S2ORC (Semantic Scholar Research Corpus)~\cite{lo2019s2orc}, which includes a large collection comprising of approximately 81.1 million English-language academic papers from various disciplines. This corpus encompasses diverse information such as abstracts, full paper texts, bibliographic references, and associated metadata for each paper.

To focus specifically on the field of Computer Science, we filtered the corpus by selecting papers with the "Field of Study" tag as Computer Science. Out of the total 81.1 million papers, we narrowed it down to approximately 6.0 million papers relevant to the computer science domain. However, not all papers in this subset had valid abstracts and body text. Some contained irrelevant or empty content, which we subsequently removed during the data cleaning process. As a result, our final dataset comprised approximately 1,00,000 samples.
\begin{table*}[ht]
\caption{Dataset statistics that we extracted from the S2ORC corpus.}\label{tab:dataset}
    \centering
    \setlength{\tabcolsep}{8\tabcolsep}
    \begin{tabular}{l|c|c|c|c}
     \hline
     \hline
     \textbf{Statistic} & \textbf{CTG-S2ORC} & \textbf{Train} & \textbf{Validation} & \textbf{Test}\\
     \hline
     \# citations & 1,00,000 & 79,588 & 9,944 & 9,946\\
     \hline
     \# unique papers & 38,530 & 34,147 & 8,076 & 8,070\\
     \hline
     \multicolumn{5}{c}{\textbf{CITATIONS}} \\
     \hline
     Avg \# characters & 171.3 & 171.34 & 171.47 & 170.87\\
     \hline
     Max \# characters & 3420 & 2986 & 1840 & 2170\\
     \hline
     \multicolumn{5}{c}{\textbf{SOURCE ABSTRACTS}} \\
     \hline
     Avg \# characters & 1225.89 & 1185.74 & 1189.3 & 1190.75\\
     \hline
     Max \# characters & 56771 & 8945 & 7858 & 8024\\
     \hline
     \multicolumn{5}{c}{\textbf{TARGET ABSTRACTS}} \\
     \hline
     Avg \# characters & 1065.64 & 1002.27 & 1001.54 & 993\\
     \hline
     Max \# characters & 93551 & 8059 & 7736 & 6647\\
     \hline
     \hline
    \end{tabular}
\end{table*}

The "body\_text" of each paper consisted of paragraphs, including sections such as Introduction, Methodology, Conclusion, etc. Within these sections, we examined the presence of cite\_spans, which are dictionaries containing citation information for referenced papers within each paragraph. Our analysis involved identifying these cite\_spans within the body text and extracting the corresponding citation sentences. It is important to note that we excluded citations that referenced more than one paper within a single sentence. Additional statistical details about the dataset can be found in Table.~\ref{tab:dataset}.

\section{\textbf{Experiments}\label{experiments}}
We describe our experimental settings, evaluation metrics, and model comparisons in this section. We fine-tuned and evaluated three Large Language text generation models on our CTG dataset. Table~\ref{tab:results1} presents the results. Comparing LLaMA~\cite{LLaMA}, Alpaca~\cite{alpaca}, and Vicuna~\cite{vicuna2023}, we observed Vicuna's superior performance. 

Next, we integrate knowledge graphs constructed from the source and target papers using PL-Marker~\cite{ye2022plmarker} and proceed to fine-tune the same set of models. The incorporation of knowledge graphs significantly enhances both the performance and the quality of the generated text. Notably, the Alpaca exhibits superior performance, as evidenced by notable increase in METEOR score by 33.14\%, and 36.98\% in Rouge-1. The Table.~\ref{tab:results2} shows the outcomes for this configuration. These results affirm that the inclusion of knowledge graphs effectively guides the Large Language Models (LLMs) in text generation tasks. 

\subsection{\textbf{Experimental Settings}}
In this research, we partitioned our CTG dataset into 79,588 training samples and 9,946 testing samples and 9,944 validation samples. For fine-tuning the Large Language Models (LLMs), we employed QLora~\cite{dettmers2023qlora} to minimize GPU usage. By back propagating gradients through a frozen, 4-bit quantized pretrained language model into Low Rank Adapters (LoRA), QLora is an effective method for maximizing memory utilization. We utilized the AdamW optimizer~\cite{kingma2014adam} with a Linear Scheduler. The learning rate was set to 3e-4, and we incorporated 100 warmup steps to gradually adjust the learning rate. By adopting this approach, we effectively trained the LLMs on the CTG dataset, allowing us to evaluate their performance on the respective testing samples.
\begin{equation}
    k_i = \frac{1}{2}\left(Q_X\left(\frac{i}{2^n + 1}\right) + Q_X\left(\frac{i+1}{2^n + 1}\right)\right)
\end{equation}

Where, $Q_x (.)$ is the quantile function of the standard normal distribution $N(0, 1)$. For our experiments, we have used $n = 4$ as we are applying 4-bit quantization.
\begin{table}[ht]
\caption{Results of CTG Task using LLMs (Without Knowledge Graphs)}
\label{tab:results1}
\vspace{0.2cm}
    \centering
    \setlength{\tabcolsep}{10\tabcolsep}
    % \resizebox{\linewidth}{!}{
    \begin{tabular}{c|c|c|c|c}
     \hline
     \hline
     \textbf{Model} & \textbf{METEOR} & \textbf{Rouge-1} & \textbf{Rouge-2} & \textbf{Rouge-L} \\
     \hline
     \textbf{LLaMA} & 12.83 & 11.26 & 1.36 & 9.59 \\
     \hline
     \textbf{Alpaca} & 10.53 & 9.22 & 1.21 & 7.81 \\
     \hline
     \textbf{Vicuna} & \textbf{14.15} & \textbf{12.88} & \textbf{1.52} & \textbf{10.94} \\
     \hline
     \hline
     \end{tabular}
    % }
\end{table}

\textbf{Evaluation Metrics:} For the text generation and summarization tasks, we employed common evaluation metrics as METEOR, ROUGE-N, and ROUGE-L. ROUGE-L evaluates the longest common subsequence between the text generated and the reference, while the overlap of n-grams between the two is measured using ROUGE-N. ROUGE-N receives additional information from METEOR, which takes word similarity into account during stemming.
\begin{table}[ht]
\caption{Results of CTG Task using LLMs (With Knowledge Graphs)}
\label{tab:results2}
\vspace{0.2cm}
    \centering
    \setlength{\tabcolsep}{10\tabcolsep}
    \begin{tabular}{c|c|c|c|c}
     \hline
     \hline
     \textbf{Model} & \textbf{METEOR} & \textbf{Rouge-1} & \textbf{Rouge-2} & \textbf{Rouge-L} \\
     \hline
     \textbf{LLaMA} & 11.61 & 10.61 & 0.99 & 9.01 \\
     \hline
     \textbf{Alpaca} & \textbf{14.02} & 12.63 & \textbf{1.54} & 10.71 \\
     \hline
     \textbf{Vicuna} & 13.80 & \textbf{12.87} & 1.48 & \textbf{10.96} \\
     \hline
     \hline
     \end{tabular}
\end{table}
\section{\textbf{Evaluations}\label{evaluations}}
Our proposed work highlights the use case of Large Language Models (LLMs) in the domain of generating citation text for scientific papers. Furthermore, our study emphasizes the significance of knowledge graphs generated from both the source and target papers, as they facilitate capturing deeper relationships and structured contextual data between these papers. Through our research, we have showcased the effectiveness of the Alpaca LLM in generating citation text, which outperforms LLaMA and Vicuna in terms of both the results obtained and the quality of the generated text. Fig.~\ref{generated_example} and Fig.~\ref{generated_example2} illustrate citation examples generated by the best model during the inference process, providing compelling evidence of the exceptional quality achieved. These findings underscore the value of leveraging LLMs and incorporating knowledge graphs to enhance the generation of accurate and contextually appropriate citation text for scientific papers.

\section{\textbf{Conclusion}\label{conclusion}}
This paper explores the task of generating citation texts in research papers. To accurately understand and capture relevant features from scientific papers, we leverage the synthesis of knowledge graphs. We present a compelling use case for employing Large Language Models (LLMs) in the domain of citation text generation, demonstrating their impressive performance when given the source abstract and target abstract, introduction, and conclusion. The efficiency of LLMs is substantiated through automatic evaluations employing various metrics. Our experiments also emphasize the significance of utilizing knowledge graphs as prompts to guide the model's generation process. Looking ahead, we plan to further enhance the capabilities of LLM models by incorporating Chain-of-Thoughts prompting, which will improve their reasoning abilities and enable the generation of more plausible and higher-quality citations.

\section{\textbf{Limitations}\label{limitations}}
While our proposed solution excels in generating single-sentence citations, its effectiveness is primarily constrained in scenarios where authors frequently employ multiple citations within a single paragraph. To overcome this limitation, we can enhance our model by incorporating multi-citation examples into our dataset. 

Another limitation of our proposed work is the presence of certain keywords in the target citation text that are not found within the section constrained by the token limit of the source and target papers. This discrepancy adversely affects the performance of the models, resulting in a decrease in overall effectiveness.

\section{Acknowledgements}
Rajiv Ratn Shah is partly supported by the Infosys Center for AI, the Center for Design and New Media, and the Center of Excellence in Healthcare at IIIT Delhi.

\bibliographystyle{splncs04}
\bibliography{custom}

\appendix
\section{Appendix}

This section shows the inference examples used to test the fine-tuned model and checking the generated text quality, and context.

\begin{figure*}[ht]
\centering
  \includegraphics[width=1\linewidth]{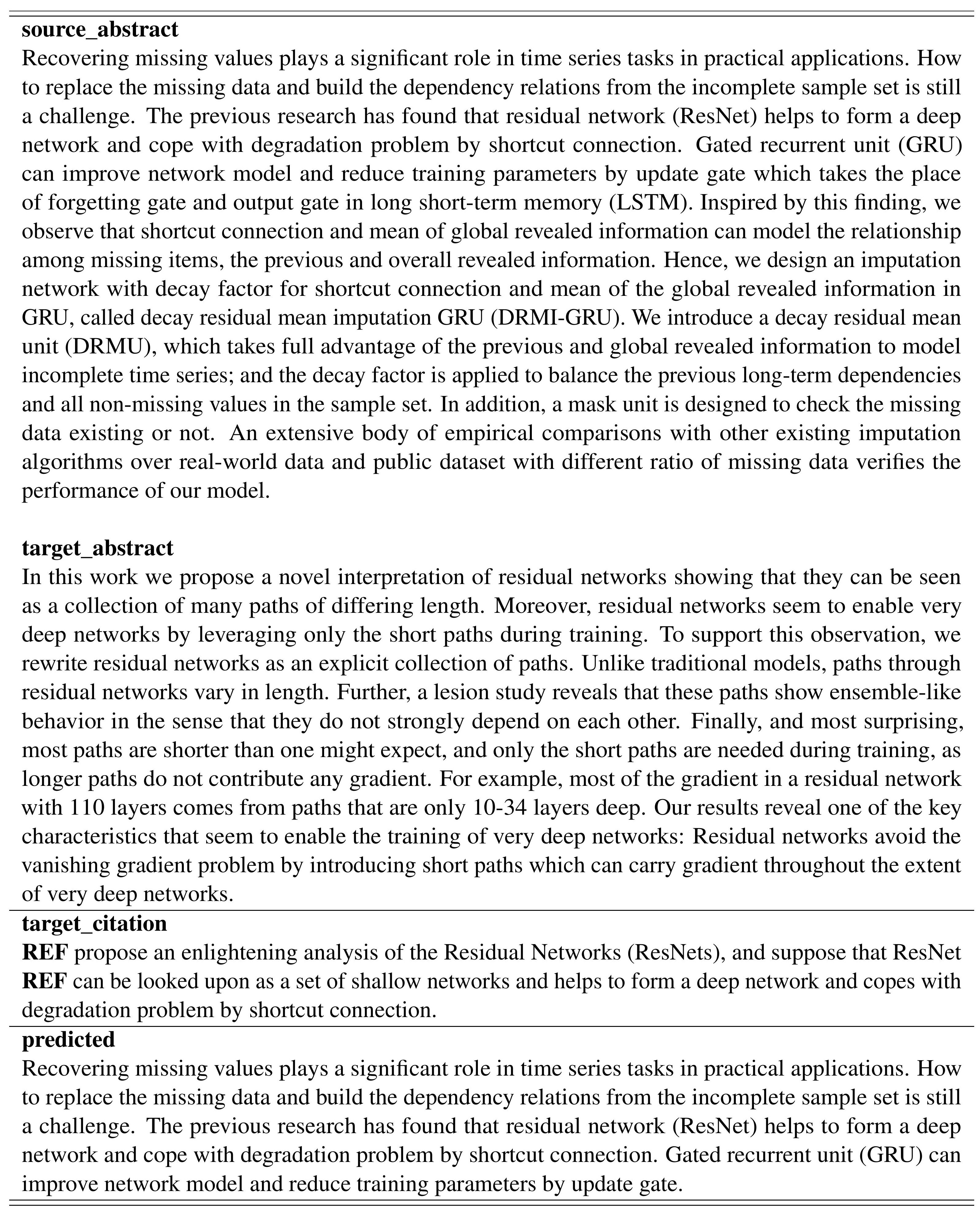}
\caption{Example of Generated Citation text from  \textbf{Vicuna} without knowledge graph relations}
\label{generated_example}
\end{figure*}

Fig.~\ref{generated_example} illustrates an example of the generated citation text obtained from the fine-tuned Vicuna model. The provided source\_abstract and target\_abstract have resulted in high-quality generated citation text that aligns well with the context of both the source and target papers.

\begin{figure*}[ht]
\centering
  \includegraphics[width=1\linewidth]{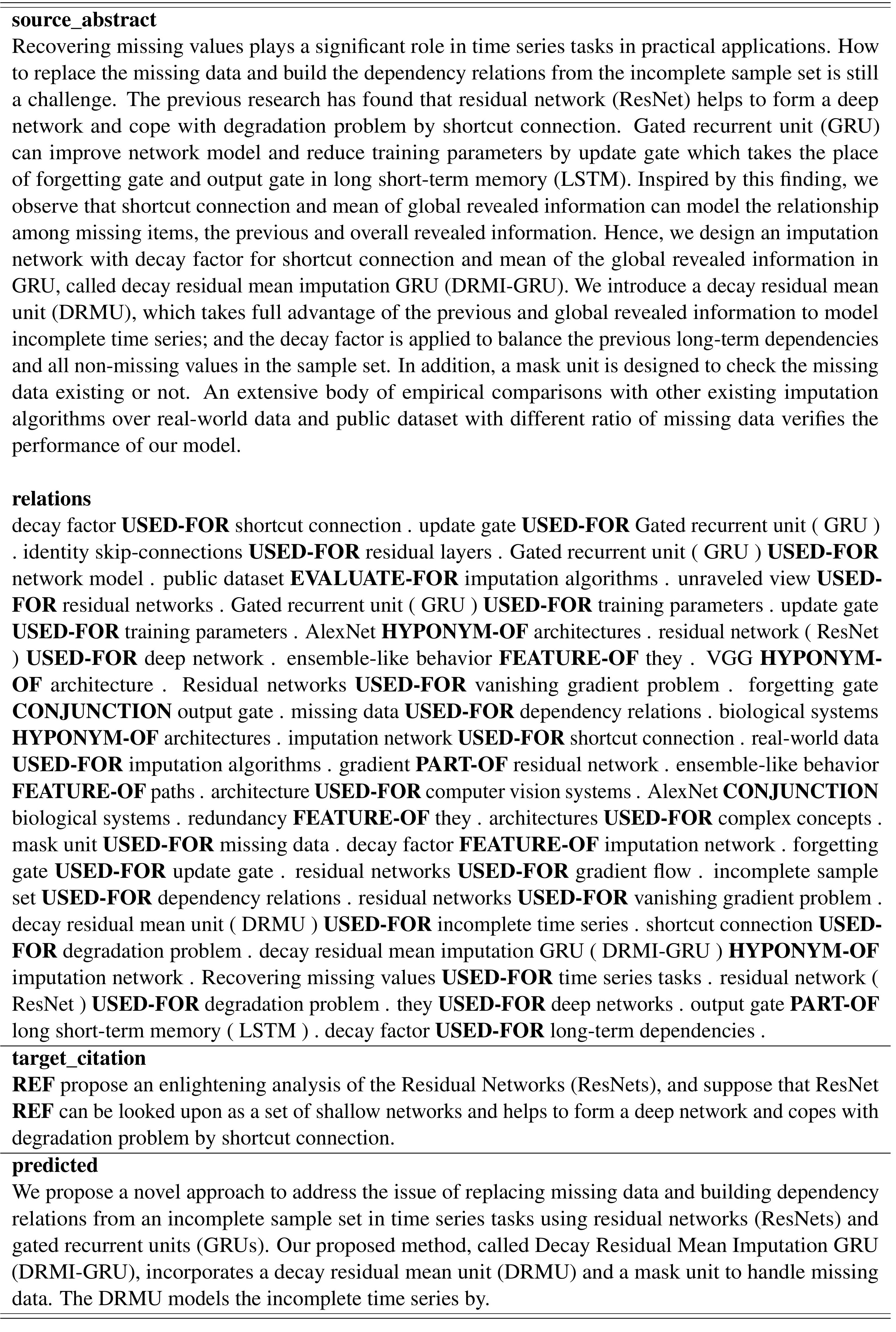}
\caption{Example of Generated Citation text from \textbf{Alpaca} with knowledge graph relations}
\label{generated_example2}
\end{figure*}

In Fig.~\ref{generated_example2}, the generated citation text demonstrates a higher level of context richness due to the incorporation of knowledge graph relations. These relations enable a better understanding of the connections between words in the source and target abstracts, resulting in more contextually relevant generated text.

\end{document}